\documentclass[letterpaper]{article} 
\usepackage{aaai25}  
\usepackage{times}  
\usepackage{helvet}  
\usepackage{courier}  
\usepackage[hyphens]{url}  
\usepackage{graphicx} 
\usepackage{natbib}  
\usepackage{caption} 
\usepackage{algorithm}
\usepackage{algorithmic}

\usepackage{newfloat}
\usepackage{listings}

\usepackage{colortbl}
\usepackage{subfigure}

\usepackage{niravstyle}
\usepackage{siunitx}
\usepackage{enumitem}
\usepackage{rotating}
\usepackage{booktabs}
\usepackage{multirow}

\newcommand{\method}{\fsc{NNRD}\xspace}

\DeclareCaptionStyle{ruled}{labelfont=normalfont,labelsep=colon,strut=off} 
\floatstyle{ruled}
\newfloat{listing}{tb}{lst}{}
\floatname{listing}{Listing}
\title{A Metric for the Balance of Information in Graph Learning}
\author {
    Alex O. Davies\textsuperscript{\rm 1},
    Nirav S. Ajmeri\textsuperscript{\rm 1},
    Telmo de Menezes e Silva Filho\textsuperscript{\rm 2}
}
\affiliations {
    \textsuperscript{\rm 1}School of Computer Science, University of Bristol \\
    \textsuperscript{\rm 2}School of Engineering Mathematics and Technology, University of Bristol\\
    alexander.davies@bristol.ac.uk, nirav.ajmeri@bristol.ac.uk, telmo.silvafilho@bristol.ac.uk
}

\usepackage{bibentry}
\begin{document}

\maketitle









\begin{abstract}
    Graph learning on molecules makes use of information from both the molecular structure and the features attached to that structure.
    Much work has been conducted on biasing either towards structure or features, with the aim that bias bolsters performance.
    Identifying which information source a dataset favours, and therefore how to approach learning that dataset, is an open issue.
    Here we propose Noise-Noise Ratio Difference (\method), a quantitative metric for whether there is more useful information in structure or features.
    By employing iterative noising on features and structure independently, leaving the other intact, \method measures the degradation of information in each.
    We employ \method over a range of molecular tasks, and show that it corresponds well to a loss of information, with intuitive results that are more expressive than simple performance aggregates.
    Our future work will focus on expanding data domains, tasks and types, as well as refining our choice of baseline model.
\end{abstract}

\section{Introduction}
\label{sec:introduction}


Graphs are an intuitive way to represent data in many fields of industry and science, including chemistry \cite{Gilmer2017NeuralChemistry}, infrastructure planning \cite{Khodayar2019DeepGrids}, biology \cite{Li2022GraphHealthcare} and social network analysis \cite{Davies2022RealisticNetworks}.
Here we focus on molecular graphs, which have seen a great deal of research and use \cite{Gomez-Bombarelli2018AutomaticMolecules, Khemchandani2020DeepGraphMolGenApproach, Popova2019MolecularRNN:Properties}.
Tasks on molecular graphs typically consist of learning to predict a property, for example solvation energy, from a set of molecular graphs.

Molecular graphs are defined by both structure and features.
Graph learning can then be framed as the process of extracting useful information from graphs for a given target.
Structure contains atoms and bonds, defining the connections between atoms within the graph.
Features then give information about individual atoms and bonds.

The degree to which Graph Neural Network (GNN) models rely on either structural or feature information during learning is highly varied.
Some works target structural learning specifically, aiming to improve performance by biasing towards the information encoded in patterns between nodes \cite{Chen2021TopologicalGraphs, Horn2021TopologicalNetworks}.
Other works make the point that the optimal solution may not include graph structure at all \cite{bechler-speicher_graph_2024}.
There are currently no metrics for whether a dataset and model combination leans towards structure or feature information, leading to complicated design processes for users.

Here we propose and demonstrate Noise-Noise Ratio Difference (\method), a metric for assessing how strongly a dataset relies on either feature-based or structure-based information, with clear applications for chemistry and other fields.
Following an intuitive noising process along either features or structure, \method gives a bounded and understandable value with which data publishers or users can tailor their approaches from the outset.

\begin{figure*}[h]
    \centering
    \includegraphics[width=\linewidth]{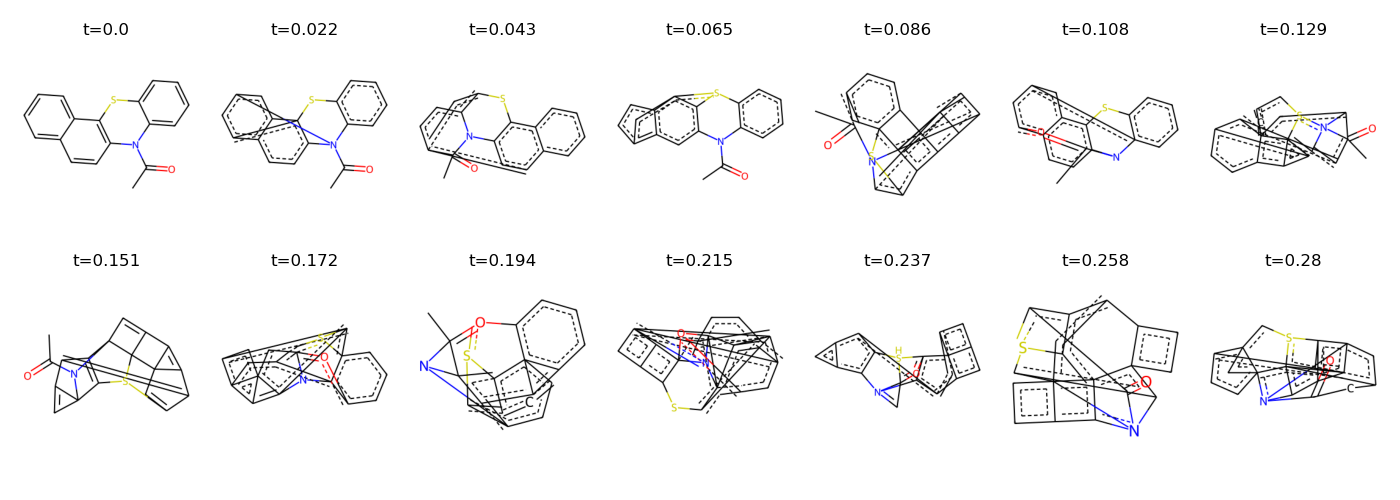}
    \caption{A molecule undergoing structure noise through edge removal and addition. Each noise step is applied on the original molecule, meaning that these examples are not sequential.}
    \label{fig:noise-structure-example}
\end{figure*}


\section{Background and Related Work} \label{sec:background}

GNNs, or more specifically Message Passing Neural Networks (MPNNs), parametrise message passing and aggregation, allowing deep-learning techniques to be applied on graph data.
For a dataset with features $X \in R^{|V| \times D}$ and similar for edges, structure as an edgelist $E:\{ (v_1, v_2), \ldots \}$, and graph labels $y$, graphs are $G:\{X, E\}$.
For a molecule, nodes are atoms and edges are bonds, with features giving information on individual atoms and bonds.
When a given model aims to learn some mapping $f(G) \rightarrow y$, in reality this is $f(X, E) \rightarrow y$.
The usual graph learning assumption is that topology $E$ and features $X$ must be considered together for information to be useful.
In other words their combination has mutual information with a target $y$, $I(y ; X, E)$.

\citet{wu_graph_2020} explore the expressive power of GNNs, introducing the information bottleneck principle.
\citet{alon_bottleneck_2021} perform similar analysis.
The core thesis of both works is that a compromise often must be struck between different information sources in graph learning.
More recent works have elaborated on the same theme, often emphasising that structural information can be difficult to effectively incorporate \cite{wu_discovering_2023}.
Other works instead show that node information can be lost during GNN use, with original feature similarities distorted during message passing and aggregation \cite{jin_node_2021}.
In a similar vein some works explicitly detach features and structure, with performance benefits, despite the loss of information from their inter-relation \cite{wang_dgnn_2024}.

Clear from the literature is that the balance between structural and feature information, and how to effectively balance the use of both, is an open issue.
While both issues are present for molecular datasets, they are also present on a diverse range of graph domains.
Here we address the former concern, aiming to provide a single metric that describes the balance between useful information sources.
Such a metric will allow chemists and other users to iterate more easily and quickly over graph learning implementations.



\begin{figure*}[ht]
    \centering
    \begin{minipage}{0.3\textwidth}
        \centering
        \includegraphics[width=\textwidth]{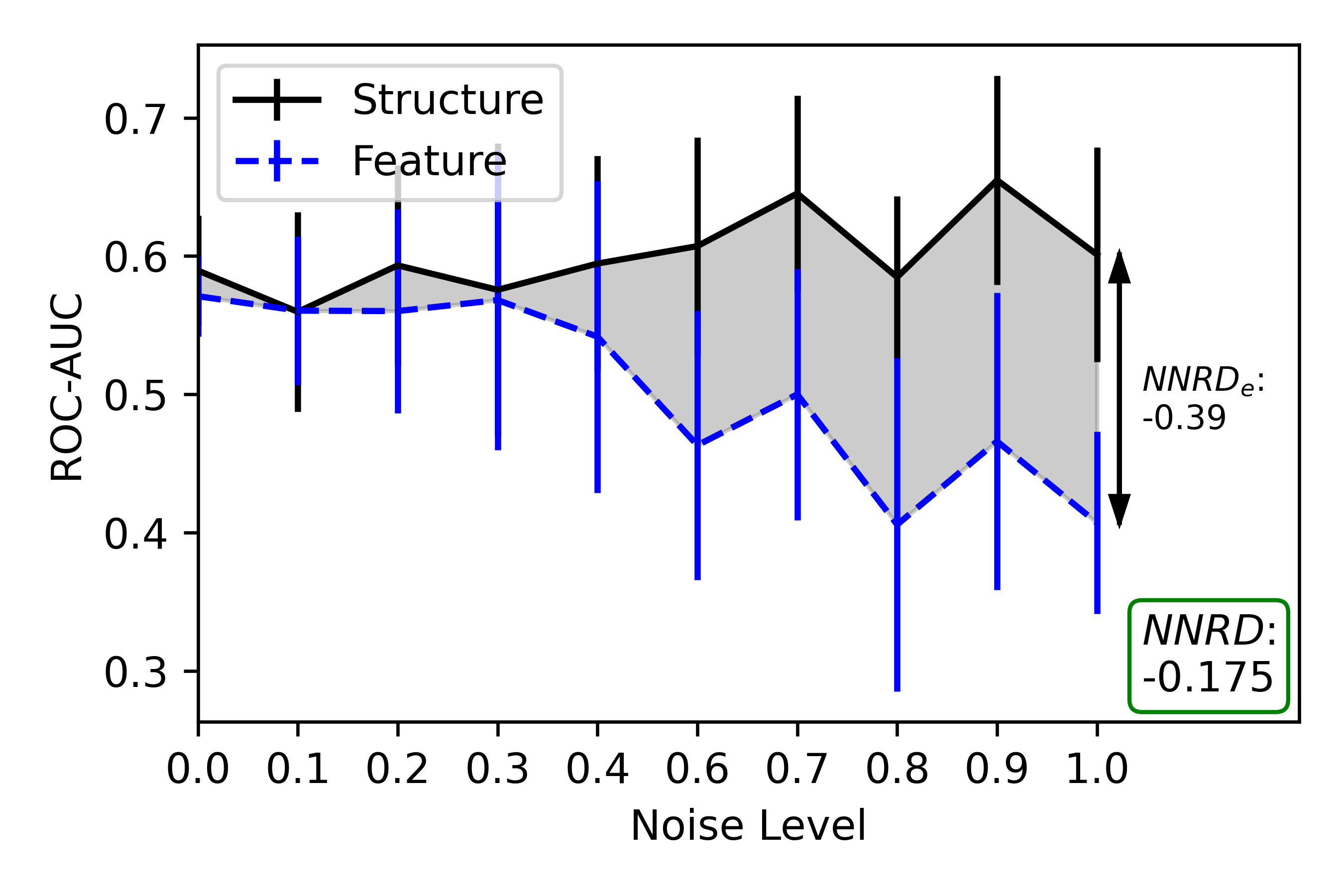} \\
        \small (a) BACE
    \end{minipage}
    \begin{minipage}{0.3\textwidth}
        \centering
        \includegraphics[width=\textwidth]{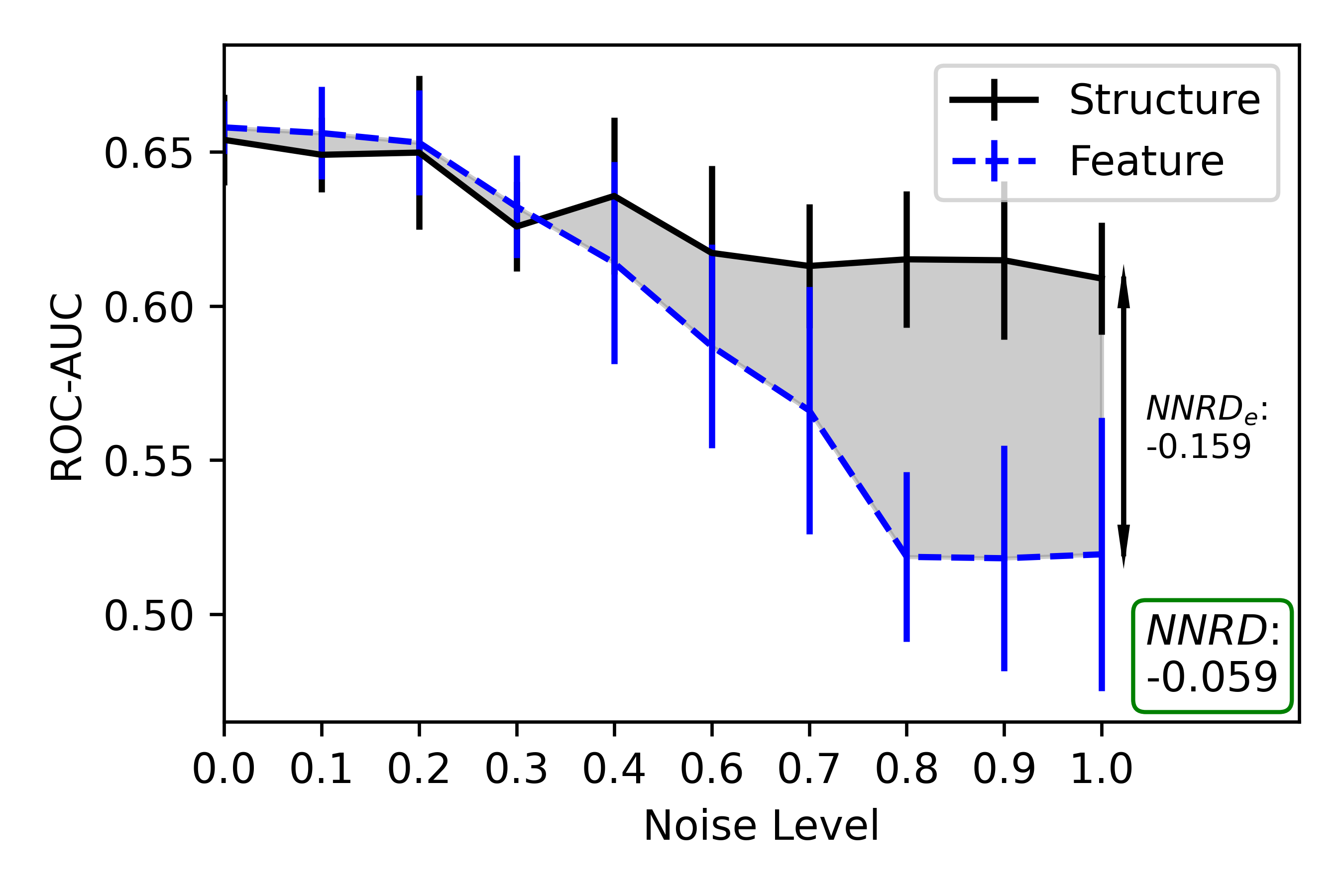} \\
        \small (b) BBBP
    \end{minipage}
    \begin{minipage}{0.3\textwidth}
        \centering
        \includegraphics[width=\textwidth]{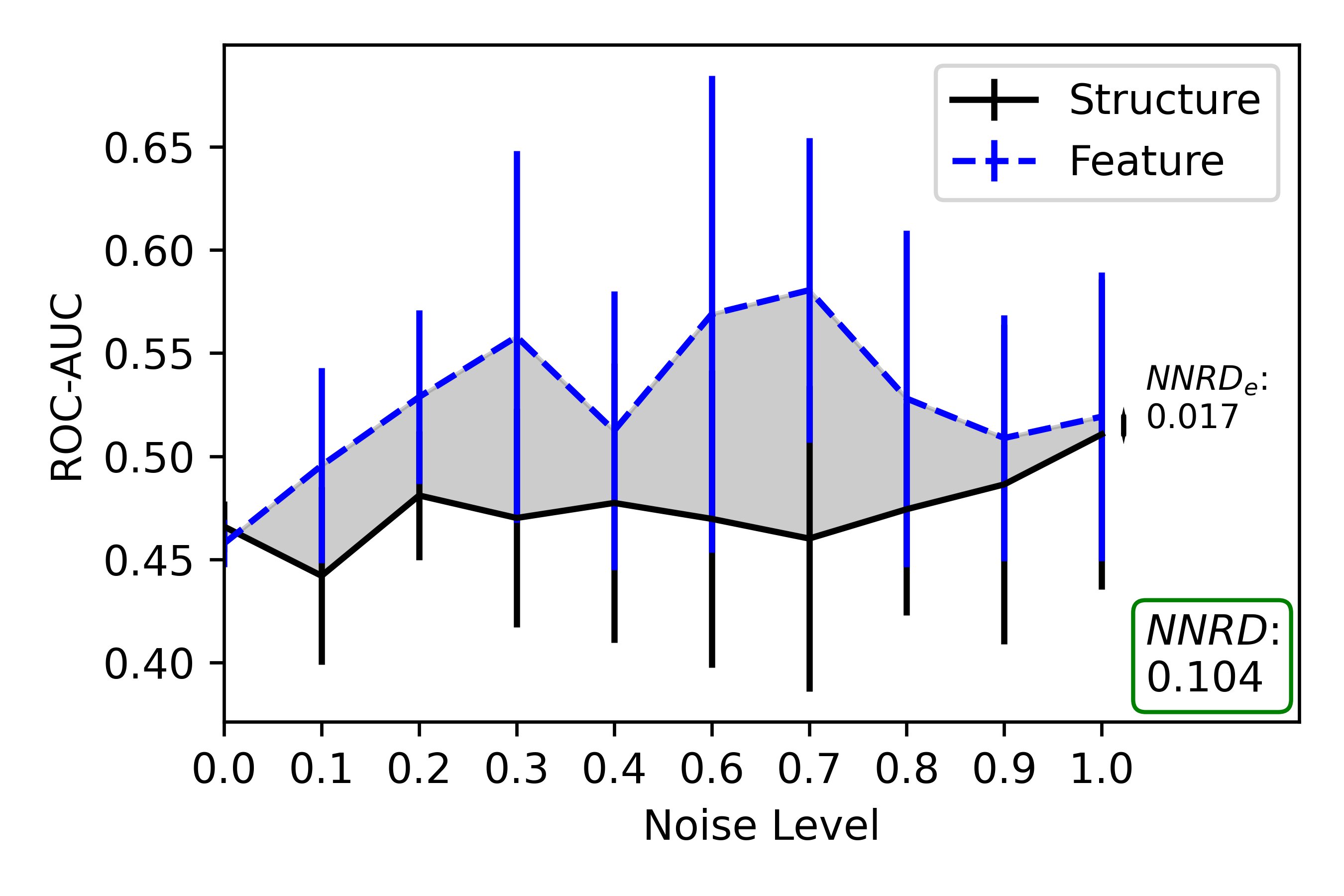} \\
        \small (c) CLINTOX
    \end{minipage} \\

    \begin{minipage}{0.3\textwidth}
        \centering
        \includegraphics[width=\textwidth]{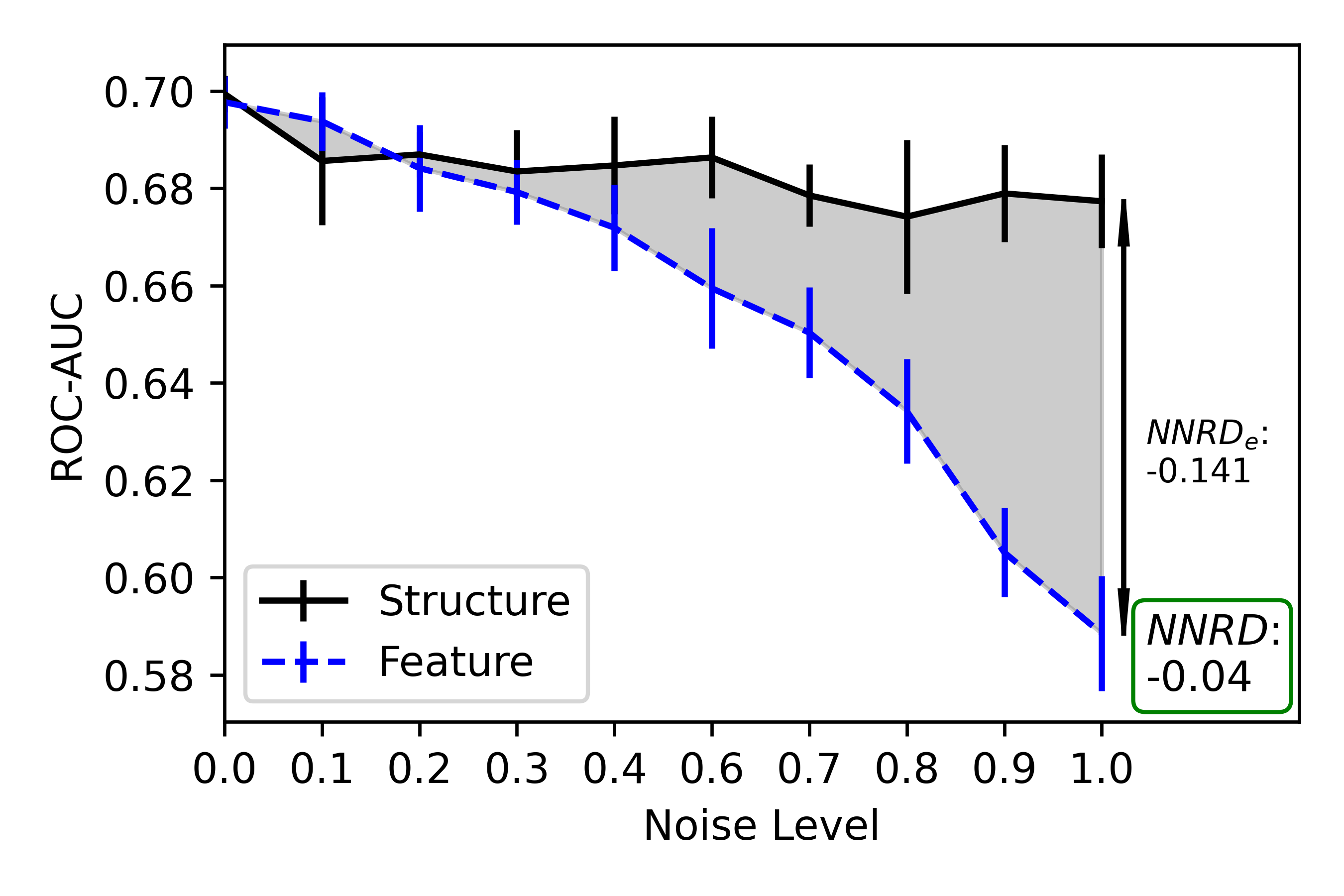} \\
        \small (d) TOX21
    \end{minipage}
    \begin{minipage}{0.3\textwidth}
        \centering
        \includegraphics[width=\textwidth]{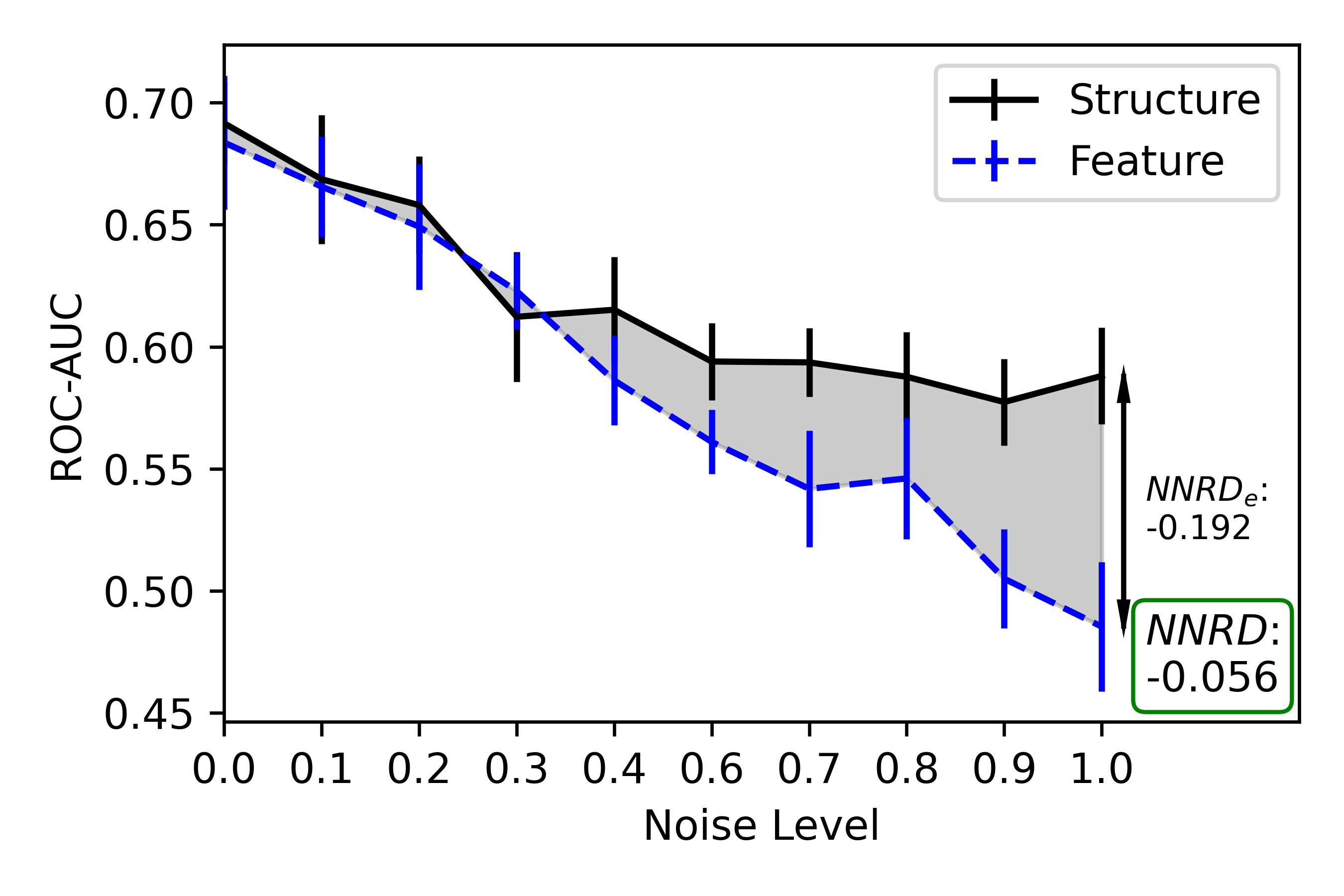} \\
        \small (e) HIV
    \end{minipage}
    \begin{minipage}{0.3\textwidth}
        \centering
        \includegraphics[width=\textwidth]{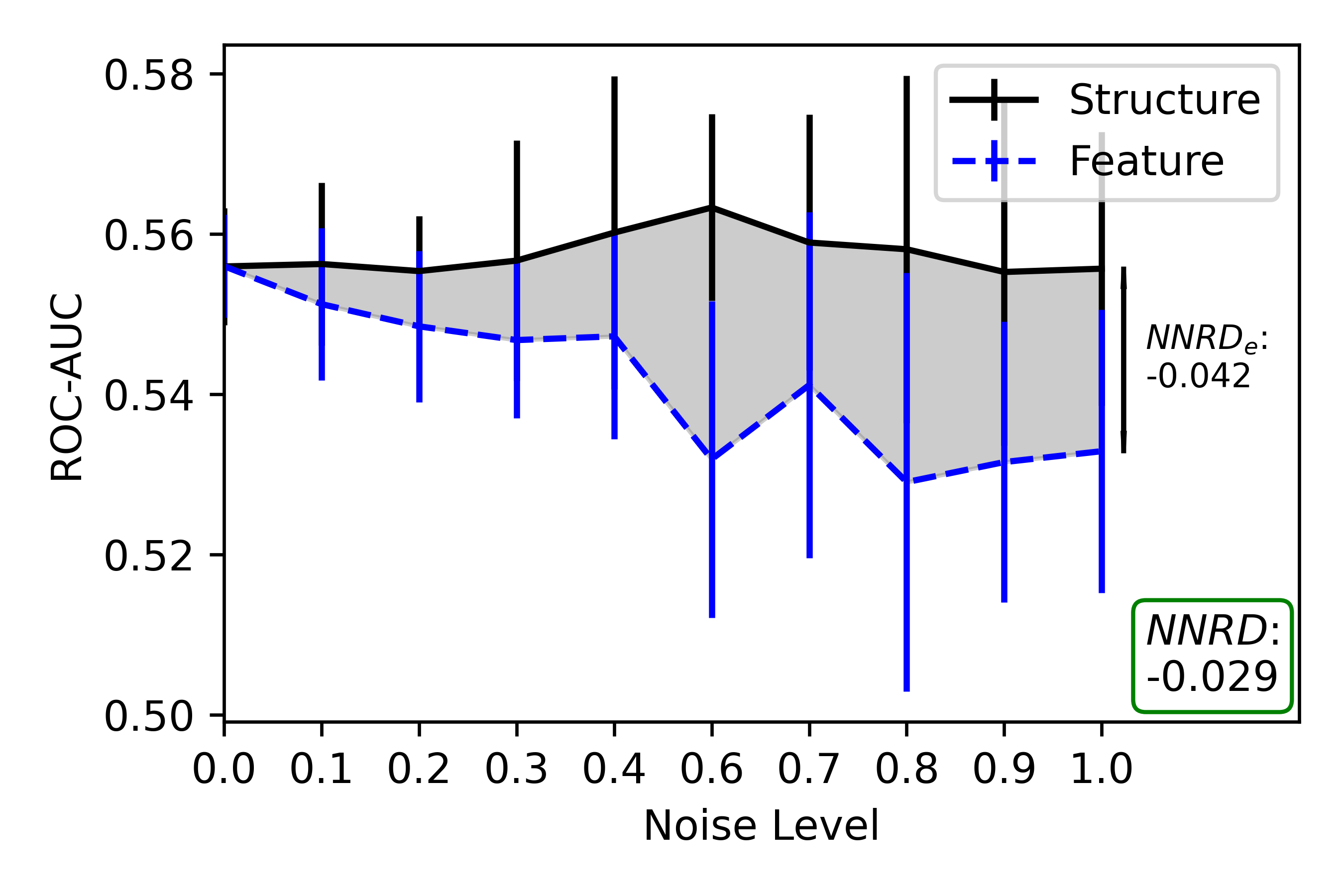} \\
        \small (f) SIDER
    \end{minipage} \\

    \begin{minipage}{0.3\textwidth}
        \centering
        \includegraphics[width=\textwidth]{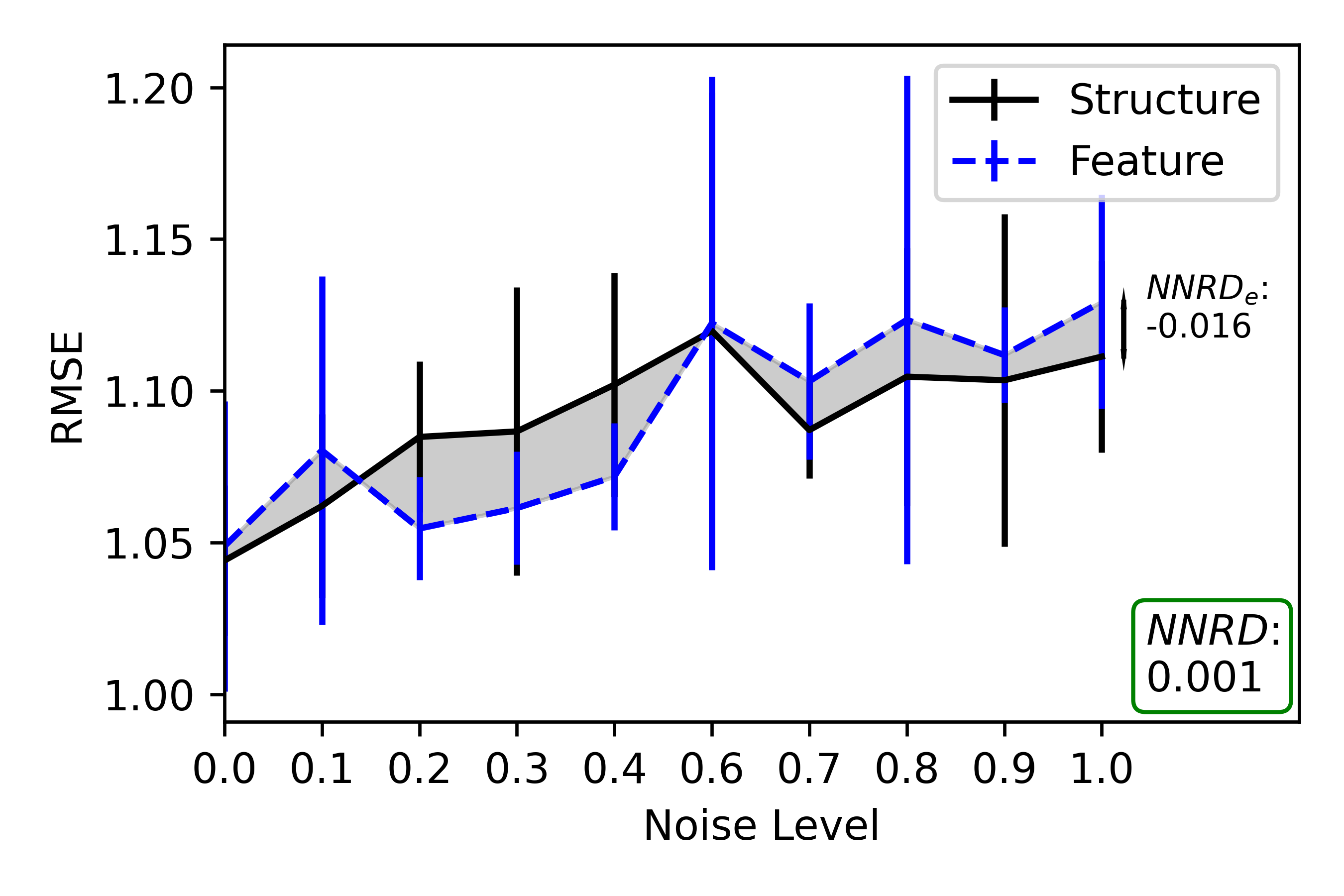} \\
        \small (g) LIPO (reg.)
    \end{minipage}
    \begin{minipage}{0.3\textwidth}
        \centering
        \includegraphics[width=\textwidth]{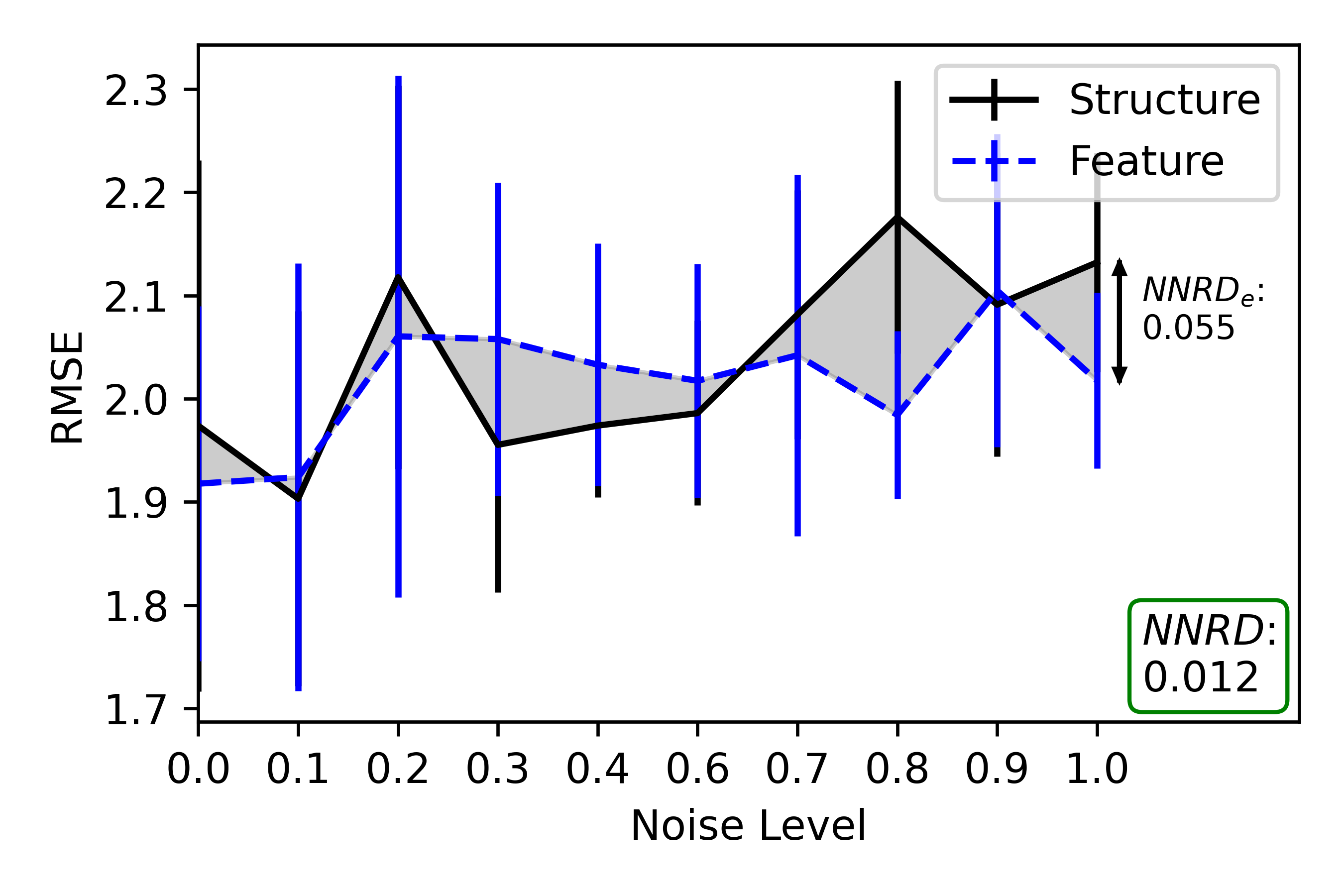} \\
        \small (h) ESOL (reg.)
    \end{minipage}
    \begin{minipage}{0.3\textwidth}
        \centering
        \small \caption{Performance variation for supervised training of our GIN models on each molecular regression benchmark dataset with increasing noise on structure and features. All datasets except LIPO and ESOL are classification, and we report ROC-AUC.}
        \label{fig:noise-noise-examples}
    \end{minipage}
    
\end{figure*}

\section{\method: Noise-Noise Ratio Difference} \label{sec:method}

We define \method here in the domain-agnostic case.
Assume that in the absence of features, structural information is still present, i.e. $I(y;E) \geq 0$, as well as that features contain their own useful information, $I(y;X) \geq 0$.
Further we can assume some imbalance between features and structure, $I(y;X) \neq I(y;E)$.

Here we propose a way of investigating this imbalance between feature information and structure information.
By degrading the useful information in either (or both of) $X$ and $E$, the degree to which performance relies on one or the other should be apparent.
A caveat here is that degrading either will presumably also degrade the information from their interdependence.

Consider some destructive noising process $N_t(\cdot)$, that degrades the useful information in either $X$ or $E$.
$N_0(x) = x$, with no useful information after $N_T(\cdot)$.
Some imperfect model produces predictions $f(X, E) \rightarrow \hat{y}$.
Performance is some $h(y, \hat{y})$ between real and predicted values.
We can then sample $h(y, f(N_X, N_E) \rightarrow \hat{y})$, the useful information in both $X$ and $E$, and whether performance requires both to be present, should be observable.

We propose Noise-Noise Difference Ratio (\method), a metric for how much a given task relies on data from features compared to structure.
Here we denote $h_X(t)$ the performance of a given model on a task with no structural noise and feature noise at some noising step $t$.
Similarly $h_E(t)$ is performance at structural noise step $t$ with no feature noise.

$t$ and $h_X(t), h_E(t)$ should monotonically increase together, as information for downstream tasks is necessarily removed by our noising functions.
If equal information is in structure and features, they will descend at the same rate.
If not, one will fall faster than the other.

From this we can define \method for performance metrics:
\begin{equation}
    \textrm{NNRD} = \log \left( \frac{1}{|T|} \sum_t \frac{h_X(t)}{h_E(t)} \right)
\end{equation}
Here we expect $h$ to increase with increasing performance.
\method is 0 when features and structure contain equal information, $>0$ when structure contains more information, and $<0$ when features contain more information.
In the same manner, when an error is tracked instead of a performance metric, the ratio term is simply inverted.
This makes direct comparison between datasets with different performance targets, for example ROC-AUC and RMSE, possible.

\section{Experiments}

We measure \method over the graph-level tasks from the Open Graph Benchmark (OGB) \cite{Hu2020OpenGraphs}.
These are a mix of binary classification, multi-class binary classification, and single-target regression datasets.
As some multi-task datasets have missing values for different tasks, we train and report metrics by masking task-wise.

We use three Graph Isomorphism Network (GIN) layers \cite{Xu2019HowNetworks} as our model, given their guarantees on expressivity, with a hidden dimension of 100 and trained with a learning rate of $0.001$.
We train each model and noised dataset five times, measuring the mean performance across ten increasing noise levels.

We use random edge removal/addition for structure noise, with an example shown in Figure~\ref{fig:noise-structure-example}.
At $t$, we remove $|E_r| = p_t \cdot |V|$ (with $0 \leq p_t \leq 1$) edges at random from the graph, $E' = E \setminus E_r$.
We then add an equal number of randomly chosen edges $|E_a| = p_t \cdot |V|$, for a final noised edgelist $E'' = E' \cup E_a$.
Where edge features are present, they are transferred from removed edges onto newly added edges.
This is very close to the procedure taken by graph diffusion models as forward noising processes \cite{Vignac2023DiGress:Generation}.
At $t = T$, $p_t = 1$, and the structure is a random graph with the same density as the original.


For feature noise we employ random feature permutation.
At $t$, we completely permute the features of a proportion $p_t$ of nodes across the dataset.
Let node (or edge) features across a whole dataset be $X \in \{0,1\}^{N \times D}$ with $N$ points and $D$ features.
At $t$ we randomly select $N_f = p_t \cdot N$ node to permute.
The features of the selected nodes are then randomly permuted, across the whole dataset, removing the useful feature information from the selected nodes, while maintaining marginal distributions.

\begin{table}[h]
    \centering
    \caption{Results from training over molecular datasets.}
    \label{tab:results}

    \begin{tabular}{ccccc}\toprule
     \multirow{2}{*}{Dataset} & \multirow{2}{*}{t=0} & \multicolumn{2}{c}{$t=T$} & \multirow{2}{*}{\method} \\
                              &                      & Structure   & Feature     &                         \\\midrule
     bace                     & 0.575                & 0.601       & 0.407       & -0.175 \\
     bbbp                     & 0.658                & 0.609       & 0.520       & -0.059 \\
     clintox                  & 0.466                & 0.511       & 0.519       &  0.104 \\
     tox21                    & 0.700                & 0.677       & 0.589       & -0.040 \\
     hiv                      & 0.684                & 0.588       & 0.485       & -0.056 \\
     sider                    & 0.556                & 0.556       & 0.533       & -0.029 \\
     lipo                     & 1.05                 & 1.11        & 1.13        & 0.001 \\
     esol                     & 1.94                 & 2.13        & 2.02        & 0.012 \\
     \bottomrule
    \end{tabular}
\end{table}

\subsection{Results}

We show visualisations of increasing structure and feature noise, and its influence on model performance, in Figure~\ref{fig:noise-noise-examples}.
Here we show NNRD, and also NNRD$_e$, which is calculated only at the extreme noise level $t=T$.
Here we see a variety of trends, varying by datasets.
On the bbbp and tox21 datasets there is a reliance on feature-based information, with scores dropping a large margin as feature noise increases, but structure noise having little effect.

On other datasets, such as the esol dataset, performance seems to decrease at an equal rate when structure and feature noise increase.
On the bace dataset, interestingly, the performance of the model actually increases with structure noise.
This may be due to additional long-range connections from edge swapping.

Notably, only on the clintox dataset do we see a preference for structural information over feature information.
Oddly performance actually increases as more noise is applied to features.
Performance over structure noises is close to constant, as on other datasets.
The likely conclusion here is that the model is simply unable to learn the dataset usefully enough to accurately predict the test set.

In Table~\ref{tab:results} we record quantitative metrics.
Here the same trends as discussed from Figure~\ref{fig:noise-noise-examples} are present, although the simple aggregate metrics we report are of a lower fidelity than the visualisations.
Crucially, in some cases, these aggregates are be misleading.

Taking the esol dataset as an example, the performance achieved at maximum noise values would indicate that structural information plays a more significant part than feature information.
However, from Figure~\ref{fig:noise-noise-examples} we can see that in-fact they increase at very similar rates, indicating a near-balance in useful information contained by each.
\method  instead reports a value very close to zero, indicating near-parity between the information in structure and features.

\section{Discussion}

While \method corresponds with the visualisations in Figure~\ref{fig:noise-noise-examples}, there are some cases where it fails.
The critical example here is the clintox dataset, where the test set deviates significantly from the train set, and performance on the test set is consistently worse-than-random as a result.
That said, this dataset is constructed as a challenging benchmark, and on other datasets \method is a useful quantitative measure.

An obvious point is that \method, in its current form, varies on a per-model basis.
This is a difficult obstacle to overcome, in that there is not currently a de-facto best model to use across all datasets.
Instead, by fixing the model used, \method could be reported on a per-model basis.
That said, the same trends would likely be present with other models.

We envision use-cases mainly in data publishing.
At the time of publishing, \method can be reported for a few fixed models, giving an indication to users of whether to design for structure or feature information.

\section{Conclusion \& Future Work}

We have proposed a metric Noise-Noise Ratio Difference (\method ) to evaluate whether the useful information on a given graph task weighs towards structure or features.
By implementing additive noise functions over structure and features, and making use of OGB datasets,  we have demonstrated the efficacy of \method.
Our future work will follow several key directions:

\begin{description}
    \item[Wider range of datasets] By including datasets from a wider range of domains, types and tasks, instead of only molecule graphs here, we can establish whether \method is also useful outside of chemistry.
    \item[Model selection] We plan to refine and tailor the models we use to calculate \method. Though a one-size-fits-all method is unlikely, a selection of different graph model approaches - for example linear models with random kernels - might provide adequate coverage.
\end{description}

\end{document}